\documentclass[journal, twoside]{IEEEtran}

\usepackage{graphicx} 
\usepackage[cmex10]{amsmath} 
\usepackage{amssymb}  
\usepackage[caption=false,font=footnotesize]{subfig}
\usepackage{algorithm}
\usepackage{algpseudocode}
\usepackage{multicol}
\usepackage{minibox}
\usepackage{color}
\usepackage{cite}
\usepackage{pstool}
\usepackage{psfrag}

\DeclareMathAlphabet{\mathtst}{OMS}{cmsy}{m}{n}

\DeclareFontEncoding{FML}{}{}%
\DeclareFontSubstitution{FML}{futm}{m}{it}%
\DeclareSymbolFont{fourier}{FML}{futm}{m}{it}%
\DeclareMathSymbol{\partialup}{\mathord}{fourier}{130} 

\newcommand{\kinect}[0]{Kinect$\textsuperscript{\textregistered}$}

\usepackage{ifthen,version}
\newboolean{include-notes}
\setboolean{include-notes}{true}
\newcommand{\tmnote}[1]{\ifthenelse{\boolean{include-notes}}%
  {\textbf{(TM says: #1)}}{}}
\newcommand{\tmadd}[1]{\ifthenelse{\boolean{include-notes}}%
  {\textcolor{red}{#1}}{#1}}
\newcommand{\jsnote}[1]{\ifthenelse{\boolean{include-notes}}%
  {\textbf{(JS says: #1)}}{}}
\newcommand{\jsadd}[1]{\ifthenelse{\boolean{include-notes}}%
  {\textcolor{red}{#1}}{#1}}

\begin{document}
%
\title{Trajectory Synthesis for Fisher Information Maximization}
%
%
%

\author{Andrew D. Wilson,~\IEEEmembership{Student Member,~IEEE,}~Jarvis A. Schultz,~\IEEEmembership{Student Member,~IEEE,}\linebreak
        and~Todd D. Murphey,~\IEEEmembership{Member,~IEEE}
\thanks{A. Wilson, J. Schultz, and T. Murphey are with the Department of Mechanical Engineering,
        Northwestern University, 2145 Sheridan Road, Evanston, IL 60208, USA. 
        awilson@u.northwestern.edu, jschultz@u.northwestern.edu, and t-murphey@northwestern.edu}%
\thanks{This is the accepted preprint version of the manuscript.  The final published version is available at http://dx.doi.org/10.1109/TRO.2014.2345918.}}%

\markboth{IEEE TRANSACTIONS ON ROBOTICS}%
{Wilson \MakeLowercase{\textit{et al.}}: Trajectory Synthesis for Fisher Information Maximization}
%

\maketitle

\begin{abstract}
Estimation of model parameters in a dynamic system can be significantly improved with the choice of experimental trajectory.  For general, nonlinear dynamic systems, finding globally ``best'' trajectories is typically not feasible; however, given an initial estimate of the model parameters and an initial trajectory, we present a continuous-time optimization method that produces a locally optimal trajectory for parameter estimation in the presence of measurement noise.  The optimization algorithm is formulated to find system trajectories that improve a norm on the Fisher information matrix.  A double-pendulum cart apparatus is used to numerically and experimentally validate this technique. In simulation, the optimized trajectory increases the minimum eigenvalue of the Fisher information matrix by three orders of magnitude compared to the initial trajectory.  Experimental results show that this optimized trajectory translates to an order of magnitude improvement in the parameter estimate error in practice.
\end{abstract}

\begin{IEEEkeywords}
optimal control, parameter estimation, maximum likelihood estimation
\end{IEEEkeywords}

\section{Introduction}
%
%
%
%
\IEEEPARstart{T}{he} design of trajectories for experimental identification of parameters in dynamic systems is an important problem in a variety of fields ranging from robotics to biology to chemistry and beyond.  Improving model accuracy facilitates the design of quality controllers which can greatly increase system performance.  When attempting to estimate a set of system parameters from observable data, the choice of experimental trajectory can have a significant effect on the precision of the parameter estimation algorithm; however, for nonlinear dynamic systems, the trajectory is constrained to nonlinear equations of motion which leads to challenges in parameter estimation and trajectory optimization.  Additionally, since the trajectories evolve on a continuous-time domain, it is important to ensure that the dynamics are satisfied throughout the entire time domain of any synthesized trajectory. 

A variety of estimation techniques are used in practice including Kalman filtering, maximum likelihood estimators, and Monte-Carlo estimators \cite{Kalman1961,Ljung1999,Bar-Shalom2004}.  This paper will focus solely on the problem of estimating static model parameters in nonlinear dynamic systems.  A widely used method for estimating static model parameters is a maximum-likelihood estimator known as the batch least-squares estimator \cite{Bar-Shalom2004}.  The batch least-squares method compares a set of measurements taken along the evolution of a trajectory to predicted observations of the system using the model equations and estimates of the parameters.  This comparison is made using the sum of the least-squares error along the trajectory from which a new update to the parameter estimate is then calculated.

A fundamental quantity that affects how well the batch least-squares estimator performs with respect to parameter estimation is Fisher information.  Finding a trajectory that maximizes the Fisher information of the model parameters will produce the best experimental estimate of the parameters, as indicated by the Cramer-Rao bound \cite{Rao1947}.  With these notions in mind, we will introduce a continuous-time optimization method that locally maximizes the Fisher information for a given nonlinear dynamic system.

\subsection{Related Work}
Since the design of an experimental trajectory has a wide range of potential uses, there have been a number of contributions to the area from different fields.  A large amount of literature on optimal experimental design exists in the fields of biology \cite{Faller2003,Baltes1994,FelixOliverLindner2006}, chemistry \cite{Franceschini2008}, and systems \cite{Dietrich2010,Martensson2011,Joshi2009,Gevers2011}.  Many of these results focus on particular applications to experiments specific to their respective fields; however, the underlying principles of information theory remain the same.

A common metric used in these areas of experimental design---also the key metric in this paper---is the Fisher information matrix computed from observations of the system trajectory \cite{Lehmann1998}.   Metrics on the Fisher information are used as a cost function in many optimization problems including work by Swevers on ``exciting'' trajectories \cite{Swevers1997}.  This work, as well as related works \cite{Armstrong1989,Gautier1992}, synthesize trajectories for nonlinear systems that can be recast as linear systems with respect to the parameters.

Further research has resulted in optimal design methods for general nonlinear systems.  In work by Emery \cite{Emery1998},  least-squares and maximum-likelihood estimation techniques are combined with Fisher information to optimize the experimental trajectories.  In this case and a number of others, the dynamics are solved as a discretized, constrained optimization problem \cite{Mehra1974,Vincent2010}.

Discretization of the dynamics \emph{a priori} has several problems.  First and foremost, an arbitrary choice about the discretization needs to be made.  Secondly, adaptive time-stepping methods cannot be used, and a discretization appropriate for the initial trajectory cannot be expected to be appropriate for the final trajectory.  Lastly, discretization can lead to high-dimensional constrained optimizations (dimensions of $10^7$ to $10^{12}$ are common in practical problems) that are impractical to solve numerically.  
To avoid discretization of continuous dynamics, a class of methods has been developed that relies on sets of basis functions to synthesize optimal controls for the system \cite{Wahlberg2010,Rackl2012,Wu2012}.  These methods allow the full trajectory to be optimized on a continuous-time domain; however, the optimization problem is still subject to a finite set of basis function coefficients.  One example of the basis function set includes the Fourier basis, which is used create a class of trajectories over which the optimization can be performed \cite{Park2006}.

In the robotics field, trajectory optimization and parameter identification algorithms have been developed  for special classes of robotic systems.  For serial robot arms and similarly connected systems, chain-based techniques and linear separation of parameters can be used \cite{Gautier1990,Radkhah2007}.  Techniques have also been adapted for parallel robots and manipulators \cite{Vivas2003,Kopacek2009,Farhat2008}.  While these techniques perform well for the intended class of robots, we seek an algorithm that has the ability to work on general nonlinear systems, only requiring differentiability of the dynamics and some form of control authority.

\subsection{Contribution}
The main contribution of this paper is the formulation of a continuous-time trajectory optimization algorithm that maximizes the information obtained from discrete observations during execution of a trajectory.  The algorithmic approach we use is based on projection-based optimal trajectory tracking \cite{Hauser2002,Hauser1998}; however, this paper extends the algorithm to include a non-Bolza cost function---maximizing the information gained by observations of the dynamic trajectory.  The extension of this algorithm is related to prior work by the authors on optimizing trajectories to improve estimation conditioning and convergence rates \cite{Wilson2013} with this paper presenting a derivation and results using Fisher information optimization to improve the accuracy of estimation.  A preliminary version of this contribution without experimental validation was presented at the 2014 American Control Conference \cite{WilsonACC}.

The benefit of using a variational technique is that a perturbation to the trajectory can be formulated in the continuous-time domain without the use of basis functions to define the trajectory.  This allows for adaptive timestepping when numerically solving the differential equations governing the system dynamics.  As an example, for the first iteration of the cart-pendulum example presented in Section V, the average timestep used to compute the trajectory perturbation is 0.0051 seconds with a standard deviation of 0.0029 seconds; however, the minimum timestep used over the trajectory is $3.6\cdot10^{-7}$ seconds.  An algorithm discretizing the dynamics at a fixed timestep  equal to this minimum timestep produces an optimization problem with $\sim n\cdot10^{7}$ dimensions, making the calculation impractical.  Subsequent iterations have varying timesteps with the minimum timestep of the final iteration at $7.4\cdot10^{-6}$ seconds.  Hence, optimizing control in a finite-dimensional, fixed timestep setting may yield intractable optimization problems compared to adaptive timestep techniques. 

This paper is organized as follows: Section II introduces the problem formulation and estimation concepts required to compute the Fisher information matrix for the system;  Section III derives the required cost function and optimization routine for the optimization over the Fisher information;  Section IV presents an overview of the experimental system that is used to validate our optimization routine; and Section V presents the experimental results.  Detailed equations for the calculation of the descent direction are provided in Appendix A.

\section{Problem Formulation}
This paper considers estimation of a set of parameters in a system  assumed to have noisy measurements but negligible process noise.  This assumption will allow the expression for Fisher information to be simplified later in the paper into a closed form.  Output variables may be a function of the states, controls, and parameter set.  The noise from measurements is assumed to be independent and normally distributed. Thus, the model of the system is defined as 
\begin{align}
\dot{x}(t) =& f(x(t),u(t),\theta)
\label{eq:dynamics}\\
y(t)=&g(x(t),u(t),\theta)+\mathrm{w}_y\nonumber
\end{align}
where $x\in\mathbb{R}^n$ defines the system states, $y\in\mathbb{R}^h$ defines the measured outputs, $u\in\mathbb{R}^m$ defines the inputs to the system, $\theta\in\mathbb{R}^p$ defines the set of model parameters to be estimated, and $\mathrm{w}_y$ is additive output noise where $p(\mathrm{w}_y)=N(0,\Sigma)$. 

\subsection{Least-Squares Parameter Estimation}
Since normally distributed measurement noise is assumed on $y(t)$, a least-squares estimator is used to estimate the system parameters.  This method is equivalent to maximum likelihood estimation due to the assumption of Gaussian noise \cite{Bar-Shalom2004}.  Using a set of measurements, nonlinear batch least-squares estimation can be performed using either a gradient descent or Newton-Raphson search method.

The least-squares estimator can be written as
\begin{eqnarray}
\hat{\theta} = \arg\min_{\theta} \beta(\theta) \label{eq-parameterestimate}
\end{eqnarray}
where
\begin{equation}
\label{eq:leastsquares}
\beta(\theta)=\frac{1}{2}\sum\limits_{i}^{h}(\tilde{y}(t_i)-y(t_i))^T\cdot\Sigma^{-1}\cdot(\tilde{y}(t_i)-y(t_i)).
\end{equation}
$\tilde{y}(t_i)$ is the observed state at the $i^{th}$ index of $h$ measurements, $\Sigma\in\mathbb{R}^{h\times h}$ is the covariance matrix associated with the sensor measurement error, and $\hat{\theta}$ is the least-squares estimate of the parameter set.

Given this estimator, we will use Newton's method with a backtracking linesearch to find optimal parameter values by minimizing the least-squares error in (\ref{eq:leastsquares}).  To perform this optimization, the first and second derivative of $\beta(\theta)$ w.r.t. $\theta$ must be calculated.

The first derivative can be computed by the following equation:\footnote{The notation, $D_\kappa\alpha(\kappa)$ represents the partial derivative of $\alpha$ w.r.t $\kappa$.}

\begin{equation}
D_\theta \beta(\theta) = \sum\limits_{i}^{h}(\tilde{y}(t_i)-y(t_i))^T\cdot\Sigma^{-1}\cdot \Gamma_\theta(t_i)
\label{firstderiv}
\end{equation}
where
\begin{align*}
\Gamma_\theta(t_i)=&D_xg(x(t_i),u(t_i),\theta) \cdot D_\theta x(x(t_i),u(t_i),\theta)\nonumber\\
&+D_\theta g(x(t_i),u(t_i),\theta).
\end{align*}

This equation requires the evaluation of $D_\theta x(x(t_i),u(t_i),\theta)$ of (1), which is computed by the following ordinary differential equation (ODE): 
\begin{equation}
\dot{\psi}(t) = D_x f(x(t),u(t),\theta)\cdot \psi(t)+D_\theta f(x(t),u(t),\theta),
\label{eq:gradient}
\end{equation}
where
\begin{equation*}
\psi(t)=D_\theta x(x(t),u(t),\theta) \in \mathbb{R}^{n\times p}\hspace{0.1in} \mathrm{and}\hspace{0.1in} \psi(0)=\{0\}^{n\times p}.
\end{equation*}
The Hessian of the least-squares estimator is also required to use Newton's method.  Differentiating (\ref{firstderiv}) w.r.t. $\theta$ yields
\begin{align}
\label{secondderiv}
& D^2_\theta \beta(\theta) = \sum\limits_{i}^{h}\Gamma_\theta(t_i)^T\cdot\Sigma^{-1}\cdot\Gamma_\theta(t_i)+(\tilde{y}(t_i)-y(t_i))^T\nonumber\\
&\cdot\Sigma^{-1}\cdot \left[D_xg(\cdot)\cdot D^2_\theta x(\cdot)+D^2_x g(\cdot)\cdot \psi(t_i)+D^2_\theta g(\cdot)\right].
\end{align}
For compactness, the arguments $(x(t),u(t),\theta)$ are replaced by $(\cdot)$.  Lastly, the formulation for the Hessian requires $D^2_\theta x(\cdot)$, which is given by the following ODE:\footnote{Since the equation for $\Omega(t)$ involves tensors, we will use the notation $\left[\cdot\right]^{T_{(1,3,2)}}$ to indicate a transpose of the 2nd and 3rd tensor dimensions.}
\begin{align*}
\dot{\Omega}(t) = &\left[\left[D^2_xf(\cdot)\cdot\psi(t)+D_\theta D_xf(\cdot)\right]^{T_{(1,3,2)}}\cdot\psi(t)\right]^{T_{(1,3,2)}}\nonumber\\
&+D_xf(\cdot)\cdot\Omega(t)+D_xD_\theta f(\cdot)\cdot\psi(t)+D^2_\theta f(\cdot),
\end{align*}
where
\begin{equation*}
\Omega(t)=D^2_\theta x(\cdot) \in \mathbb{R}^{n\times p \times p} \hspace{0.1in} \mathrm{and}\hspace{0.1in} \Omega(0)=\{0\}^{n\times p\times p}.
\end{equation*}

\subsection{Estimation Algorithm}
Using (\ref{firstderiv}) and (\ref{secondderiv}), parameter estimation is performed using either gradient descent or Newton steps based on the conditioning of the optimization.  Typically, when far from an optimal solution, a few gradient descent steps are necessary before Newton steps will be effective.

Additionally, a backtracking linesearch method \cite{Armijo1966} is used to ensure that a sufficient decrease condition is satisfied for each iteration.  For a well-conditioned problem, Newton steps will generally provide quadratic convergence rates with the stepsize  $\gamma_j=1$; however, in practice, ill-conditioned problems can cause convergence rates to degrade.

The estimation algorithm is outlined in Algorithm \ref{paramalgorithm}. Iterations continue until the gradient of the cost function $\beta(\theta)$ drops to a specified tolerance level.
\begin{algorithm}
\caption{Parameter Estimation}\label{paramalgorithm}
\begin{algorithmic}[0]
\State Choose initial $\theta_0\in\mathbb{R}^p$, tolerance $\epsilon$, $j=0$
\While{$D_\theta\beta(\theta_j)>\epsilon$}
\State Calculate descent using gradient descent or Newton:
\State \hspace{0.15in}Gradient descent: $d_j = -D_\theta\beta(\theta_j)$ 
\State \hspace{0.15in}Newton: $d_j = -\left[D^2_\theta\beta(\theta_j)\right]^{-1}\cdot D_\theta\beta(\theta_j)$
\State Compute $\gamma_j$ with Armijo backtracking search
\State $\theta_{j+1}=\theta_j+\gamma_j d_j$
\State $j=j+1$
\EndWhile
\end{algorithmic}
\end{algorithm}

\subsection{Fisher Information}
Fisher information quantifies the amount of information a set of observations contains about a set of unknown parameters \cite{Lehmann1998}.   For dynamic systems, a set of measurements taken during the execution of a trajectory provides information about model parameters.  The effectiveness of the parameter estimation technique outlined in the previous section is related to the Fisher information matrix (FIM).  

Assuming that the measurement noise of the system is normally distributed with zero process noise, the FIM for the system is given by, 
\begin{equation}
\label{eq:FIM}
I(\theta)= \sum\limits_{i}^{h} \Gamma_\theta(t_i)^T\cdot \Sigma^{-1}\cdot \Gamma_\theta(t_i).
\end{equation}
Note that this expression has been simplified from the general form of Fisher information given the assumptions stated above. Details of the simplification can be found in \cite{Bar-Shalom2004}.

In practice, increasing the Fisher information of the system tends to increase the maximum precision that can be obtained by an estimation algorithm; however this is not explicitly guaranteed.  This relationship is quantified by the Cramer-Rao bound \cite{Rao1947} given by
\begin{equation}
cov_\theta(\hat{\theta}) \geq I(\theta)^{-1},
\label{eq:cramerrao}
\end{equation}
where $\hat{\theta}$ is the least-squares estimator defined in (\ref{eq-parameterestimate}).  While the Cramer-Rao bound provides the minimum possible covariance of the parameter estimate, it does not bound how large the covariance may be for a given trial.

To use the FIM as a metric for trajectory optimization, $I(\theta)\in\mathbb{R}^{p\times p}$ must be mapped to a scalar value.  There is a significant amount of literature on different types of mapping choices \cite{Faller2003,Emery1998,Mehra1974}; however, this paper will restrict itself to the design choice of E-optimality. The choice of E-optimal design is used to improve the worst-case variances of the parameter set by maximizing the minimum eigenvalue of the FIM.  Thus, the algorithm creates a variation on the trajectory to proportionally improve the information acquired.  Other experimental design approaches such as A and D-optimality conditions, which provide different goals for the desired information acquisition, can be realized with modifications to the objective function.  The following section presents the derivation of the E-optimal objective function and presents the iterative algorithm.

\section{Nonlinear Trajectory Optimization}
The trajectory optimization algorithm used in the results to follow is a projection-based optimal trajectory tracking algorithm which has been extended to include Fisher information as an objective \cite{Hauser2002,Hauser1998}. This algorithm is formulated in continuous-time to allow the time discretization of the dynamics to be easily decoupled from the measurement time points as well as allow for the use of any numerical integration method. Therefore, the objective including any costs on the Fisher information must be formulated in continuous time.
\subsection{Objective Function}
To define an objective function dependent on the FIM on a continuous-time domain, the maximization of the information matrix needs to be cast from a finite set of discrete measurements into an appropriate continuous analogue.  To satisfy this condition, the information equation (\ref{eq:FIM}) will be written as 
\begin{equation}
\tilde{I}(\theta) = \int_{t_0}^{t_f} \Gamma_\theta(t)^T\cdot \Sigma^{-1}\cdot \Gamma_\theta(t)\hspace{0.05in}dt.
\label{eq:continfo}
\end{equation}
Assuming that observations are taken regularly along the entire trajectory, the optimal trajectory $x^*(t)$ maximizing the eigenvalues of the continuous  $\tilde{I}(\theta)$ will approximately optimize the eigenvalues of the sampled $I(\theta)$.  As the sampling rate increases, the values of $\tilde{I}(\theta)$ and $I(\theta)$ will converge.  

If measurement only occur along certain portions of the trajectory with predetermined times, a weighting function can be added to $\tilde{I}(\theta)$ to ensure that sensitivity is maximized specifically in the sampled areas of the trajectory.  However, if the observation time is not predetermined, the sensitivity along the entire trajectory will be maximized using $\tilde{I}(\theta)$.

The optimization objective function will therefore be given by
\begin{align}
J=\frac{Q_p}{\lambda_{min}}+\frac{1}{2}\int\limits_{t_0}^{t_f}&\left[(x(t)-x_d(t))^T\cdot Q_\tau \cdot (x(t)-x_d(t))\right.\nonumber\\
&\left.+u(t)^T\cdot R_\tau \cdot u(t)\right] dt
\label{eq:trajcostcont}
\end{align} 
where $\lambda_{min}$ is the minimum eigenvalue of $\tilde{I}(\theta)$, $Q_p$ is the information weight, $x_d(t)$ is a reference trajectory, $Q_\tau$ is a trajectory tracking weighting matrix, and $R_\tau$ is a control effort weighting matrix. The weights must be chosen such that $Q_p\geq 0$, $Q_{\tau}$ is positive semi-definite, and $R_{\tau}$ is positive definite.

The various weights allow for design choices in the optimal trajectory that is obtained.  The requirements of positive definiteness and positive semi-definiteness of the weighting matrices are necessary to maintain a locally convex optimization problem including the fact that $\lambda\geq 0$ \cite{Hauser1998}.  Increasing the control weight will result in less aggressive trajectories, generally decreasing the obtained information.  Using a reference trajectory allows for an optimal solution that remains in the neighborhood of a known trajectory.  

While the cost function is generally well defined, it does become ill-conditioned and singular in the case of a zero eigenvalue.  This will occur if the initially chosen trajectory yields no information about at least one unknown parameter.  For a dynamic system, this may occur if a stationary trajectory is chosen in which case, perturbing the initial trajectory of the system using a heuristic method will likely provide a sufficient initial trajectory for the algorithm.  If a perturbation fails to provide any information, the system geometry or sensor configuration may prohibit the experiment from resolving the parameters simultaneously for any trajectory.  In this case, sensors may need to be added, or unknown parameter removed to create a well-posed experiment.

\subsection{Extended Dynamic Constraints}
The optimal control algorithm was previously formulated for trajectory tracking problems where the objective function is explicitly a function of the system states \cite{Hauser1998}.  However, given (\ref{eq:trajcostcont}), the cost also depends on $\psi(t)=D_\theta x(t)$.  
Since the objective is to  minimize a norm that includes $\psi(t)$, which depends nonlinearly on $x(t)$, $\psi(t)$ is treated as an additional state.  Appending $\psi(t)$ to the state vector as an additional dynamic constraint  allows for variations in $\psi(t)$ in the optimization algorithm.
For convenience, the extended state will be defined by $\bar{x}(t)=(x(t),\psi(t))$, and $\eta(t)=(\bar{x}(t),u(t))$ defines a curve which satisfies the nonlinear system dynamics. 

\subsection{Projection Operator}
The minimization of (\ref{eq:trajcostcont}) is subject to the dynamic constraints given by (\ref{eq:dynamics}) and (\ref{eq:gradient}).  The minimization of this unconstrained optimization involves iteratively calculating a descent direction followed by a projection that maps the unconstrained trajectory, formed by the sum of the current iterate and the descent direction, onto the dynamic constraints as detailed in \cite{Hauser2002}.  The projection operator uses a stabilizing feedback law to take a feasible or infeasible trajectory, defined by $\xi(t)=(\bar{\alpha}(t),\mu(t))$, and maps it to a feasible trajectory, $\eta(t)=(\bar{x}(t),u(t))$.

The projection operator used in this paper is given by
\begin{equation*}
P(\xi(t)):\left\lbrace \begin{array}{l}
u(t)=\mu(t)+K(t)(\bar{\alpha}(t)-\bar{x}(t))  \\ 
\dot{x}(t)=f(x(t),u(t))  \\ 
\dot{\psi}(t) = D_x f(x,u,\theta)^T \psi(t)+D_\theta f(x,u,\theta)^T.
\end{array} \right.
\end{equation*}
The feedback gain $K(t)$ can be optimized as well by solving an additional linear quadratic regulation problem.  Details of the optimal gain problem can be found in \cite{Hauser2002}, but any stabilizing feedback may be used.

With the addition of the projection operator, the problem can be reformulated from an optimization over the constrained trajectory $\eta(t)$ to an optimization over the unconstrained trajectory $\xi(t)$. This relation is given by
\begin{equation*}
\arg \min_{\eta(t)} \hspace{0.05in} J(\eta(t))\hspace{0.1in} \longleftrightarrow\hspace{0.1in} \arg \min_{\xi(t)} \hspace{0.05in} J(P(\xi(t))).
\end{equation*}
The unconstrained formulation allows variations of the trajectory to be calculated free of the constraint of maintaining feasible dynamics; however, the solution is projected to a feasible trajectory at each iteration of the optimization algorithm.

\subsection{Optimization Routine}
Algorithm \ref{optalgorithm} defines the iterative method using a gradient descent approach to solve the optimization problem.   Each iteration requires a descent direction $\zeta_k(t)=(\bar{z},v)$ to be computed from the following equation \cite{Hauser1998}:
\begin{equation}
\zeta_k(t) = \arg\min_{\zeta_k(t)}DJ(P(\xi_k(t)))\circ\zeta_k(t)+\frac{1}{2}\langle\zeta_k(t),\zeta_k(t)\rangle
\label{eq:descent}
\end{equation} 
such that
\begin{equation*}
\dot{\bar{z}}=A \bar{z}+B v
\end{equation*}
where $\zeta_k(t)\in T_{\eta_k}\mathtst{T}$, i.e., the descent direction for each iteration lies in the tangent space of the trajectory manifold at $\eta_{k}$.  The components of the descent direction $\zeta_k=(\bar{z}(t),v(t))$ are defined by $\bar{z}(t)$, the perturbation to the extended state, and $v(t)$, the perturbation to the control. The quantity $\langle\zeta_k(t),\zeta_k(t)\rangle$ is a local quadratic model computed as an inner product of $\zeta_k(t)$.   Matrices $A$ and $B$ are the linearizations of the extended state dynamics.  Since (\ref{eq:descent}) is a quadratic function of $\zeta_k$ with linear constraints, the descent direction can be computed using LQR techniques which depends on the linearization of the cost function, $DJ(P(\xi_k(t)))$, and the local quadratic model, $\langle\zeta_k(t),\zeta_k(t)\rangle$.  The detailed calculation of the descent direction is provided in Appendix \ref{app:descent}.

Given the descent direction $\zeta_k$, a backtracking linesearch of the projection, $P(\eta_k(t)+\gamma_k\zeta_k)$, provides a feasible trajectory assuming the step size $\gamma_k$ satisfies the Armijo sufficient decrease condition \cite{Armijo1966}. Iterations upon the feasible trajectories continue until a given termination criteria is achieved.

\begin{algorithm}[h]
\caption{Trajectory Optimization}\label{optalgorithm}
\begin{algorithmic}[0]
\State Initialize $\eta_0\in\mathtst{T}$, tolerance $\epsilon$, $k=0$
\While{$DJ(\eta_k(t))\circ\zeta_k>\epsilon$}
\State Calculate descent, $\zeta_k$:
\State \hspace{0.15in}$\zeta_k=\arg\min_{\zeta_k(t)}DJ(P(\xi_k(t)))\circ\zeta_k+\frac{1}{2}\langle\zeta_k,\zeta_k\rangle$
\State Compute $\gamma_k$ with Armijo backtracking search
\State Calculate the infeasible step:
\State \hspace{0.15in}$\xi_{k}(t)=\eta_k(t)+\gamma_k\zeta_k$
\State Project trajectory onto dynamics constraints:
\State \hspace{0.15in}$\eta_{k+1}(t)=P(\xi_k(t))$
\State $k=k+1$
\EndWhile
\end{algorithmic}
\end{algorithm}

\section{Experimental Setup}
\begin{figure}[t]
\centering
\hspace{-0.35in}
\includegraphics{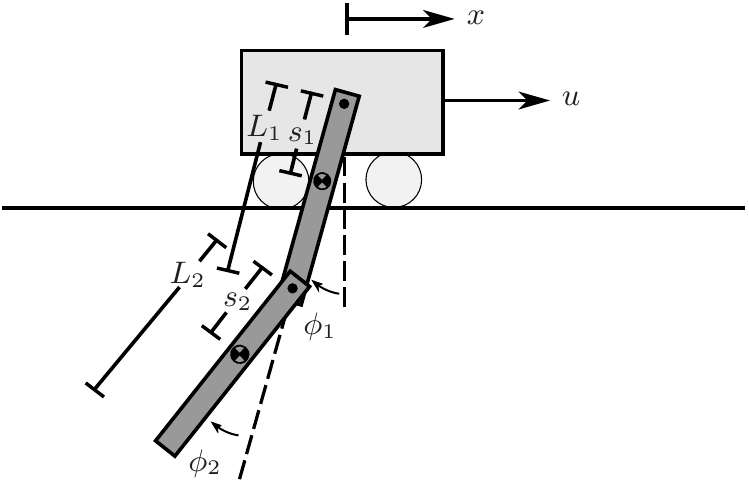}
\caption{Cart-pendulum system}
\label{fig:cart}
\end{figure}

To demonstrate the optimization of Fisher information for a dynamic system, a simulation and experimental test of a 2-link cart-pendulum system is considered.  The system has three configuration variables, $q=(x(t),\phi_1(t),\phi_2(t))$, where $x(t)$ is the horizontal displacement of the cart, and $\phi_i(t)$ is the rotational angle of each link as seen in Fig. \ref{fig:cart}. 

The control input to the system is the acceleration of the cart, given by $u$.  The cart can accelerate in either direction with positive acceleration to the right. Rotational friction is modeled at each pendulum joint, but the joints remain unactuated.  The goal of the optimization algorithm is to accurately estimate the mass of the top pendulum link and the damping coefficient of the joints.

\begin{figure}[t]
  \centering
  \def\svgwidth{3.05in}
  \input{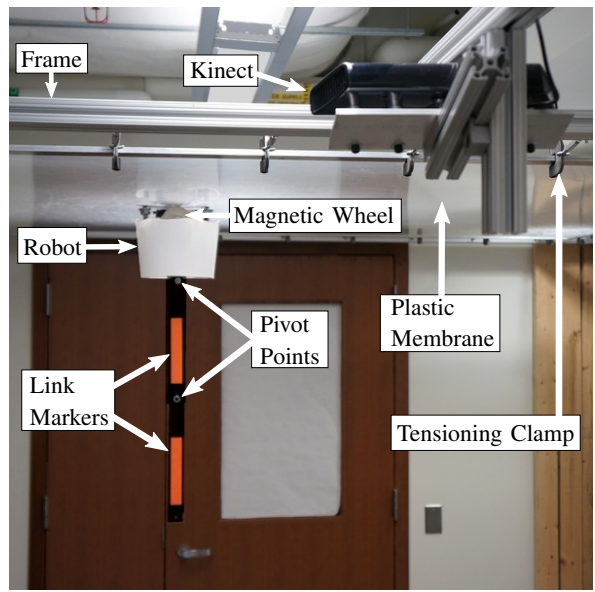}
  \caption{Image showing the experimental setup with annotations of the
    important components provided.}
  \label{fig:experiment}
\end{figure}

\subsection{Experimental Testbed Setup}
\subsubsection{Robotic System}
\label{sec:robot_system}
The experimental setup shown in Fig. \ref{fig:experiment} consists of a differential drive mobile robot with
magnetic wheels moving in a plane. The pivot point for the first link of the
double pendulum is provided by mounts on the robot.  A plastic
membrane provides the robot's driving surface.  The membrane is tensioned in all
directions within an aluminum frame mounted parallel to the ground
at a height of approximately $2.5$ meters.  An unpowered, magnetic idler
mechanism is placed on the top side of the plastic to provide adherence for the
robot.  The idler has the same geometric footprint as the robot, but its
magnetic wheels have opposite polarity.

Two 24~V DC motors drive the robot's magnetic wheels.  PID loops running at
1500~Hz close the loop around motor velocity using optical encoders for
feedback.  Two 12~V lithium iron phosphate batteries provide all required power.
Desired velocity commands are sent to the robot wirelessly using Digi
XBee$\textsuperscript{\textregistered}$ modules at 50~Hz, and a 32-bit Microchip
PIC microprocessor handles all on-board processing, motor control, and
communication.  The motors are powerful relative to the inertias of the robot,
the double-pendulum payload, and the wheels.  Thus, they are capable of
accurately tracking aggressive trajectories up to a maximum rotational
velocity. Since the input to the 2-link cart-pendulum system is the
direct acceleration of the cart, the PID loops on the motor velocity allow
the robot to accurately track a given velocity profile.  The Robot Operating
System (ROS) is used for visualization, interpreting and transmitting desired
trajectories and for processing and recording all experimental data
\cite{quigley2009ros}.

\begin{figure}[t]
  \centering
    \label{fig:track_points}
    \includegraphics[width=.7\columnwidth]{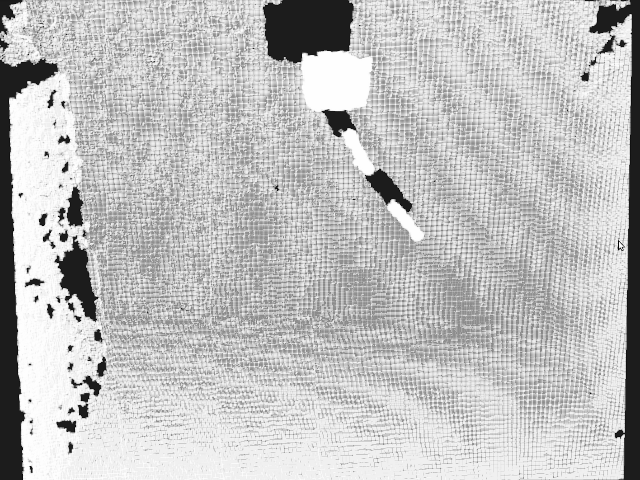}
  \\
    \label{fig:track_clusters}
    \includegraphics[width=.7\columnwidth]{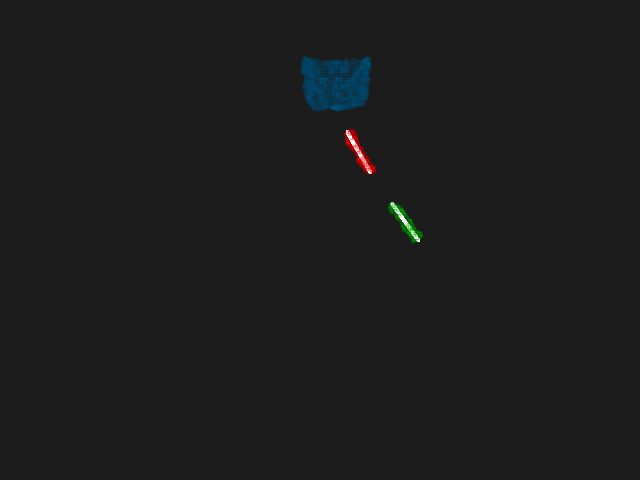}
    \caption{Illustration of the image processing performed on the point cloud
      provided by the \kinect. The top image 
      shows the raw point cloud, and the bottom 
      shows the data after all processing; different colors are assigned to each
      cluster, and white lines are displayed to show the lines produced by the
      sample consensus filter.}
  \label{fig:track}
  \vspace{-0.1in}
\end{figure}

\subsubsection{Tracking System}
A Microsoft Kinect$\textsuperscript{\textregistered}$ is employed to track the
system and obtain experimental measurements.  The Kinect provides uncolored
point clouds at 30~Hz as seen in Fig. \ref{fig:track}, and the Point Cloud Library (PCL) is used for data
processing \cite{Rusu2011}.  Each raw point cloud is first downsampled, and then
pass-through filters eliminate points that lie outside of a
predetermined bounding box of expected system configurations.  A Euclidean
cluster extraction algorithm is then used to extract three separate clouds -- one
for the robot and one for each of the links. The pendulum links are made from
an acrylic that is transparent to the Kinect; opaque markers adhered to
each of the links provide a visible surface.  The markers have a gap
between them ensuring that the software uniquely detects the three
clusters.  The coefficients of the line equations along the axis of each link
are then extracted using a sample consensus segmentation filter on the clusters
representing the links.  Once the coefficients for these lines are known, simple
trigonometry allows calculation of the desired link angles, $\phi_1$ and
$\phi_2$.

Since the Kinect will provide measurements of the absolute rotation angle of
each pendulum link, it is important to estimate the variance of the
measurements for use in the optimization algorithm.  To measure the covariance of the angles computed from the point cloud, measurements were recorded for 30 seconds at sensor's fixed frequency of 30 Hz with the double pendulum hanging in a stationary position.  Although the variance of the
measurements will have some dependence on the link velocity and configuration, we assume independence for the purpose of this experiment. The angles calculated from the point cloud are assumed to be independent for the two links, so the off-diagonal elements are set to zero, yielding a covariance matrix given by
\begin{equation}
\label{eq:covariance}
\Sigma = \left[\begin{array}{cc}
1.12\times 10^{-4}&0\\
0&4.79\cdot 10^{-4}\\
\end{array}
\right] \mathrm{rad}^2.
\end{equation}

\subsection{System Model}
A mathematical model of the system is created by deriving standard rigid body dynamics equations.  A Lagrangian,
$L = T - V$
is constructed where the kinetic energy of the system $T$ is given by 
$T=\frac{1}{2}\sum\limits_{i=1}^{2} m_i \nu_i(x,\phi_1,\phi_2)^2+\mathcal{I}_i\omega_i(x,\phi_1,\phi_2)^2$, where $\nu_i$ is the translational velocity, and $\omega_i$ is the rotational velocity of each link. The potential energy $V$ is a function of the height $z_i$ of each link's center of mass,
$V=\sum\limits_{i=1}^{2}m_i g z_i(x,\phi_1,\phi_2)$ given the gravitational constant $g$.
For each link, the moment of inertia $\mathcal{I}_i$ is defined as a function of the mass and measured center of mass location. The center of mass on the link is also used to define the potential energy of the link in the Lagrangian.

Using the Euler-Lagrange equation,
\begin{equation*}
0 =\frac{\partial L}{\partial q}-\frac{d}{dt}\left(\frac{\partial L}{\partial \dot{q}}\right),
\end{equation*}
the full equations of motion can be solved in terms of $\ddot{q}(t)$ (keeping in mind that $\ddot{x}=u$ by assumption).  

The Kinect measures the absolute angle $y_i$ of each link $i$ which provides two measurement outputs, $y(t)=[y_1(t),y_2(t)]$.  Since the system states define the angle of link 2 relative to link 1, the output function is
\begin{equation*}
g(x(t),u(t),\theta) = \left[\begin{array}{c}
\phi_1(t)\\
\phi_1(t)+\phi_2(t)
\end{array}\right].
\end{equation*}

\subsection{System Parameter Selection}
We have not yet addressed the choice of which parameters to identify for
  a given system. Choosing the set of parameters will depend on the context of
the estimation needs as well as concerns with identifiability.
In certain situations, some parameters may not be identifiable regardless
  of trajectory selection.  For example, given the 2-link cart-pendulum
system with no damping in the joints, estimating both link masses
independently is not possible.  To illustrate this, Fig. 4 shows a contour
plot of $\beta$, the parameter estimator cost, vs. changes in each mass value. The straight isolines in the plot indicate that the parameters are coupled through a constant scaling factor resulting in no unique set of mass parameters that minimize the cost.  Rather, the estimator cost can be minimized with an infinite linear combination of the mass parameters, therefore, the specific parameter set is non-identifiable by the estimator. 

The non-identifiability condition can be tested for a given trajectory by examining the eigenvalues of the Fisher information matrix.  If and only if a zero eigenvalue exists, neither parameter can be uniquely identified. If a zero eigenvalue exists for the trajectory, another trajectory should be tested for identifiability.  If no trajectory can be found, the algorithm cannot be initialized and a different set of parameters should be chosen.  The experiment that follows in this paper is performed under the assumption that the two unknown system parameters are the mass of the top link $m_1$ and the damping coefficient of the top link joint and bottom link joint $c$.  Since both links use the same bearings, we will assume that the damping coefficients of the two joints are equal.  The remaining system parameters have been measured and are presented in Table \ref{tab:measuredparam}.  Additionally, Table \ref{tab:cost} lists the eigenvalues of the FIM for the initial trajectory.  Since both eigenvalues are non-zero, the parameters can be simultaneously estimated to some precision which will be discussed in Section VI.C.

\begin{figure}[t]
\centering
\hspace{-0.35in}
\includegraphics{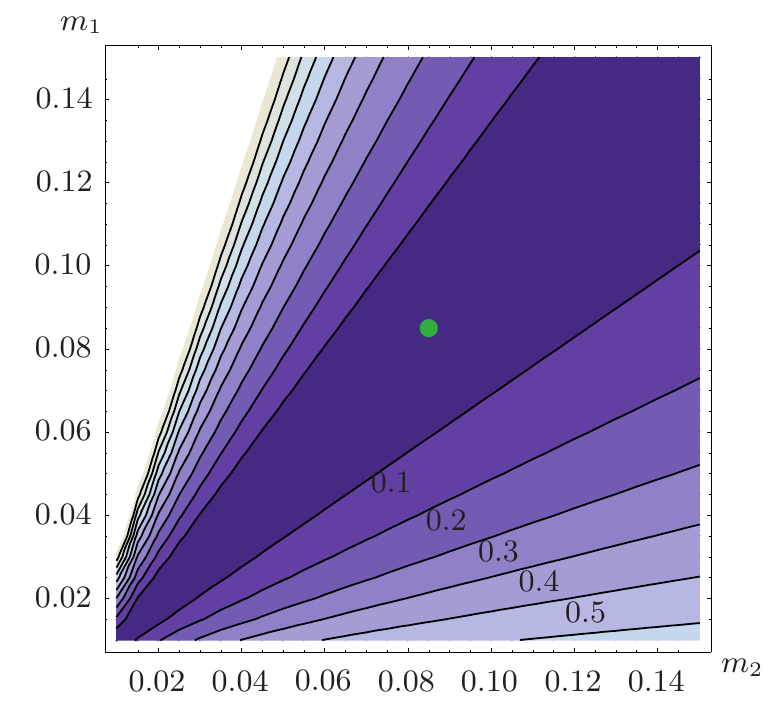}
\caption{Contour plot of unidentifiable system. The deterministic value of the parameter is shown by the green dot in the contour plot with isolines indicating the estimator cost, $\beta(\theta)$.  The straight lines indicate a poorly conditioned estimation problem which results in non-identifiable parameters.}
\label{fig:failopt}
\end{figure}

\begin{table}[t]
\caption{Measured System Parameters}
\centering
\begin{tabular}{|l c c|}
\hline
\textit{Parameter}&Link 1 Value&Link 2 Value\\
Mass, $m$&Estimate&0.0847 kg\\
Link Length, $L$&0.305 m&0.305 m\\
Link Width, $w$&0.0445 m&0.0381 m\\
Bearing offset from edge&0.0127 m&0.0127 m\\
Center of mass position, $s$&0.146 m&0.125 m\\
Damping coefficient, $c$&Estimate*&Estimate*\\
\hline
\multicolumn{3}{l}{\scriptsize*The values for the damping coefficient of each joint will be assumed to be equal.}
\end{tabular}
\label{tab:measuredparam}
\end{table}

Given that $m_1$ and $c$ are the uncertain parameters, the following values will be used as initial estimates of the parameters, 
\begin{equation}
m_1 = 0.085 \hspace{0.05in}\mathrm{kg}   \hspace{0.5in}c = 0.50 \hspace{0.05in}\mathrm{g/sec}.
\label{eq:initialestimate}
\end{equation}
Note that the choice of units weights the desired precision between estimated parameters. In this case, the damping coefficient units are scaled to provide more relative precision in the mass estimate.  The remaining system parameters which appear in Table \ref{tab:measuredparam} will be considered the actual known model parameters.

\section{Results}

\begin{figure}[t]
\centering
\subfloat[Cart position, $x(t)$]{
\includegraphics{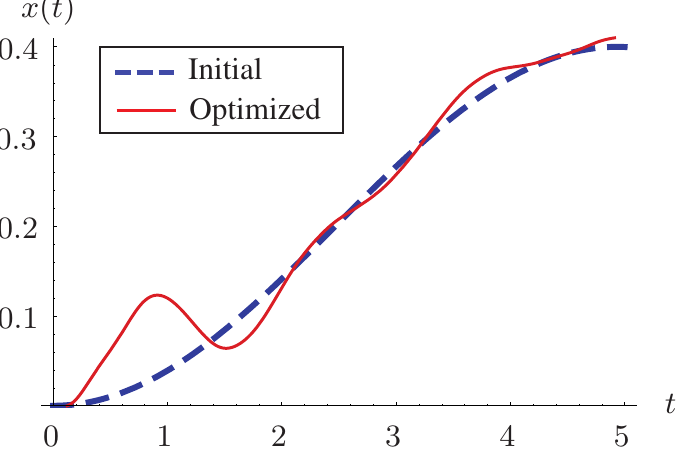}}\\
\vspace{-0.15in}
\subfloat[Link 1 angle, $\phi_1(t)$]{
\hspace{-0.35in}
\includegraphics{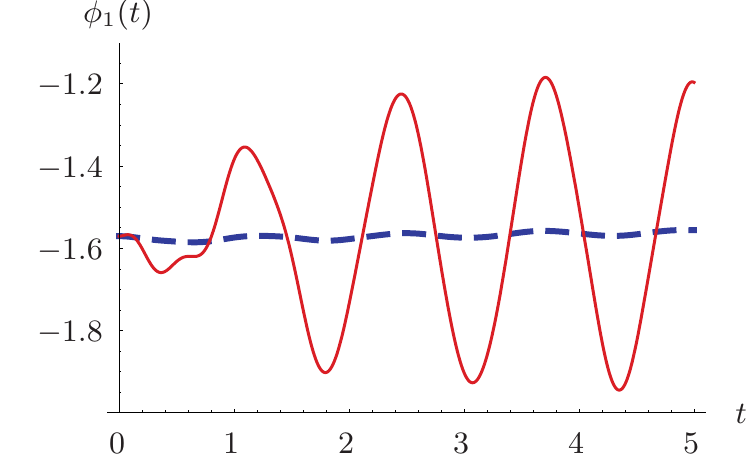}}\\
\vspace{-0.15in}
\subfloat[Link 2 angle, $\phi_2(t)$]{
\hspace{-0.2in}
\includegraphics{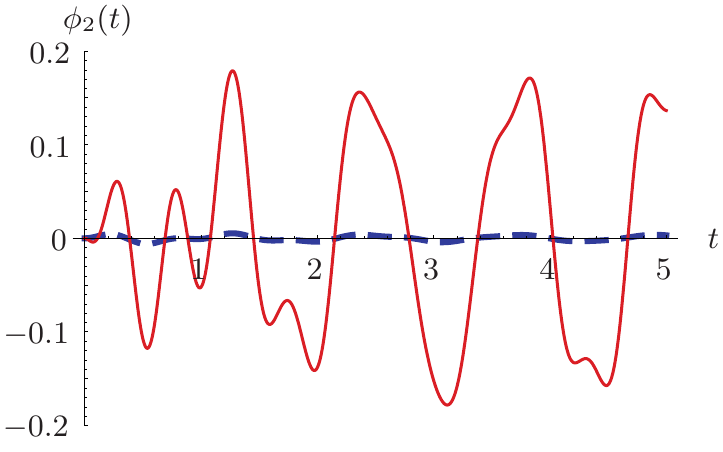}}\\
\vspace{-0.15in}
\subfloat[Control input, $u(t)$]{
\hspace{-0.23in}
\includegraphics{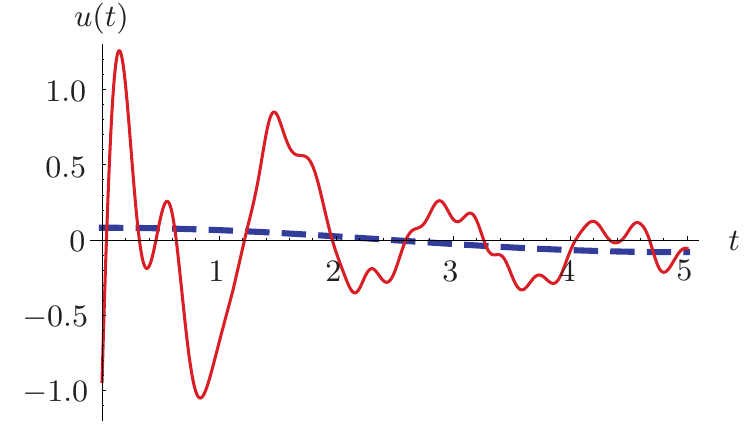}}\\
\caption{Comparisons of the trajectory before and after Fisher information optimization.}
\label{fig:opttraj}
\end{figure}

\subsection{Optimization Results}
The optimization algorithm was run until a convergence criterion of $|DJ(\xi(t))\circ\zeta|<10^{-1}$ was satisfied, starting from
an initial descent magnitude around $8\cdot10^3$.  The comparison of initial and optimized trajectories can be seen in Fig. \ref{fig:opttraj}.

Table \ref{tab:cost} shows the initial and optimized eigenvalues of the FIM, $\tilde{I}(\theta)$, and the initial and optimized cost $J$.  The results show that the minimum eigenvalue $\lambda_2$ increases by over a factor of $10^3$.  Additionally, the other eigenvalue increases, though not included directly in the cost function. 

Examining the plots of the optimized trajectory, it is clear that more information is gained by oscillating the pendulum back and forth.  In particular, more information about the damping parameter $c$ is gained by increasing the oscillation.   This observation leads to a hypothesis that the cost function and optimization may be driven strongly by information concerning the damping parameter, although the result increases the overall amount of information for both parameters.

\begin{table}[t]
\caption{Optimization Results}
\centering
\begin{tabular}{|l c c c|}
\hline
\vspace{-0.12in} &&&\\
 &$\lambda_1$&$\lambda_2$&$J$\\
Initial:&123.3&0.0315&15.9\\
Optimized:&$1.43\cdot 10^5$& $32.6$&$0.618$\\
- - - - - -& \multicolumn{3}{c|}{- - - - - - - - - - - - - - - - - - - - - - -}\\
&\multicolumn{3}{c|}{Fisher Information Matrix}\\
Initial: &\multicolumn{3}{c|}{$\left[\begin{array}{l c c}
\hspace{0.1in}3.72\cdot 10^{3}&-5.14\\
\hspace{0.2in}-5.14&\hspace{0.1in}9.57\cdot 10^{-1}\\
\end{array}\right]$}\\
&&&\\
Optimized:&\multicolumn{3}{c|}{$\left[\begin{array}{l l}
\hspace{0.1in}4.31\cdot 10^{6}&-3.77\cdot 10^{3}\\
-3.77\cdot 10^{3}&\hspace{0.1in}9.91\cdot 10^{2}\\
\end{array}\right]$}\vspace{-0.12in}\\
 &&&\\
\hline
\end{tabular}
\label{tab:cost}
\end{table}

Using this optimized trajectory, simulation and experimental results of the system as well as the experimental testbed are presented in the following sections to demonstrate the improvement in the estimation of system parameters.

\subsection{Monte-Carlo Simulation Analysis}
Given the system model and parameters, results are first obtained in simulation to assess the effectiveness of the algorithm given an exact model and pure Gaussian noise distributions.  An experimental measurement of the trajectory is simulated by generating the trajectory of the deterministic system using our estimate of the parameters and sampling the trajectory at a discrete number of points adding Gaussian additive noise at each sample.  This results in a discrete set of noisy measurements that will be used for the subsequent parameter optimization routine.  For the simulation example, a 5 second trajectory is simulated at the Kinect's sampling rate of 30 Hz.  The sampling rate is set by the frequency of sensors being used. The uncertainty of each state measurement is normally distributed with zero mean and variance equivalent to that measured by the Kinect during the testbed setup given by (\ref{eq:covariance}).

\begin{figure}[t]
\centering
\subfloat[PDF histograms of the top link mass, $m_1$.]{
\includegraphics{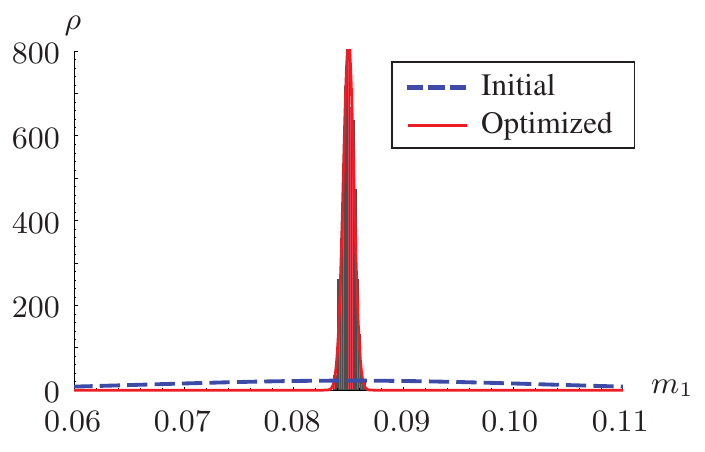}}\\
\vspace{-0.15in}
\subfloat[PDF histograms of the damping coefficient, $c$.]{
\hspace{-0.35in}
\includegraphics{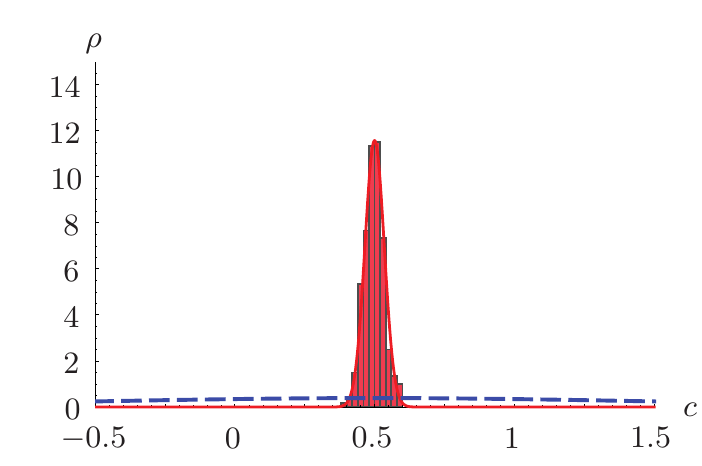}}\\
\caption{Monte-Carlo histogram approximations of the PDF with respect to the Lebesgue measure of the optimized trajectory with the optimized variance fit in red and the initial trajectory variance in blue. }
\label{fig:histogram}
\end{figure}
\begin{table}[!h]
\caption{Monte-Carlo Simulation Results}
\centering
\begin{tabular}{|c c|}
\hline
\vspace{-0.12in} &\\
&Monte-Carlo Covariance\\
Initial: &$\left[\begin{array}{l c c}
3.16\cdot 10^{-4}&3.45\cdot 10^{-3}\\
3.45\cdot 10^{-3}& 1.06\\
\end{array}\right]$\\
&\\
Optimized:&$\left[\begin{array}{l l}
2.37\cdot 10^{-7}&1.63\cdot 10^{-6}\\
1.63\cdot 10^{-6}& 1.19\cdot 10^{-3}\\
\end{array}\right]$\\
- - - - - -& - - - - - - - - - - - - - - - - - - - - - - -\\
&Cramer-Rao Lower Bound\\
Initial: &$\left[\begin{array}{l c c}
2.71\cdot 10^{-4}&1.46\cdot 10^{-3}\\
1.46\cdot 10^{-3}&1.05\\
\end{array}\right]$\\
&\\
Optimized:&$\left[\begin{array}{l l}
2.32\cdot 10^{-7}&8.83\cdot 10^{-7}\\
8.83\cdot 10^{-7}& 1.01\cdot 10^{-3}\\
\end{array}\right]$\vspace{-0.1in} \\
&\\
\hline
\end{tabular}
\label{tab:rao}
\end{table}

A Monte-Carlo simulation is performed using 300 trials, resampling the trajectory with new additive Gaussian noise samples for each trial.  For the scope of the Monte-Carlo simulation analysis, the parameter estimates from (\ref{eq:initialestimate}) are treated as the actual system parameter set used to generate the simulated output and the initial parameter estimate are uniformly varied in the range of $\pm100\%$ of each parameter value in (\ref{eq:initialestimate}).  This provides simulation results that are independent of the initial estimate.

After running the simulation, histogram plots in Fig. \ref{fig:histogram} show the distribution of parameter estimates using the optimized trajectory. The histogram of the optimized trajectory is shown by the red bins with a Gaussian fit to the mean and standard deviation shown by the solid red line.  The dashed blue line indicates the Gaussian fit to the histogram results of the initial trajectory; however, to simplify the plot, the initial histogram is not shown.   The averaged statistics of the simulation trials can be seen in Table \ref{tab:rao}.  Since the expected information has been significantly increased with the optimized trajectory, the results confirm that the covariance of the estimated parameters decreases dramatically.  The precision of both parameter estimates with respect to the Lebesgue measure is improved by a factor of $10^3$.  

\begin{figure}[t]
\centering
\subfloat[Cart position, $x(t)$]{
\includegraphics{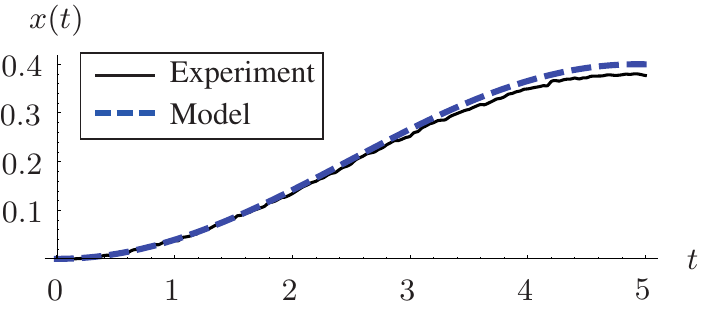}}\\
\vspace{-0.25in}
\subfloat[Link 1 angle, $\phi_1(t)$]{
\hspace{-0.35in}
\includegraphics{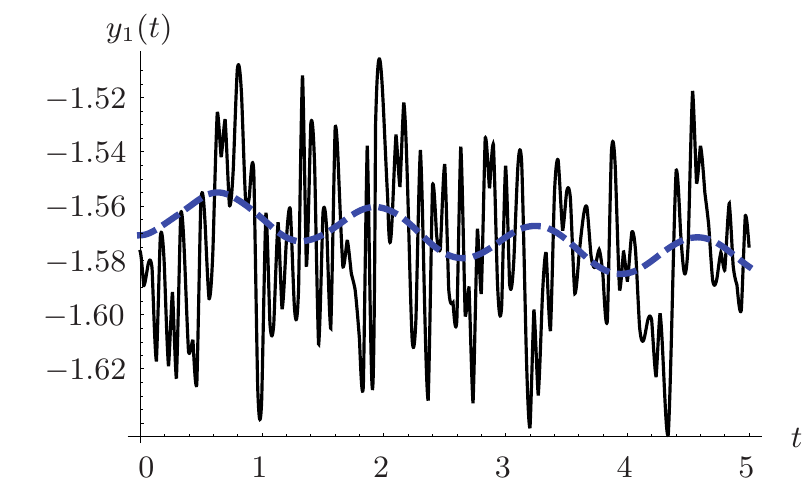}}\\
\vspace{-0.25in}
\subfloat[Link 2 angle, $\phi_2(t)$]{
\hspace{-0.35in}
\includegraphics{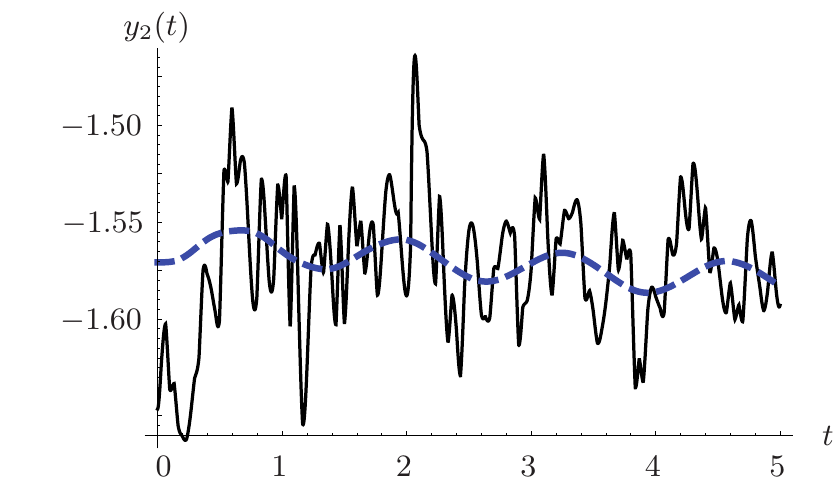}}\\
\caption{Comparisons of the measured trajectories and estimated model trajectory for the initial choice of experimental trajectory.}
\label{fig:initexp}
\end{figure}

\begin{figure}[t]
\centering
\subfloat[Cart position, $x(t)$]{
\includegraphics{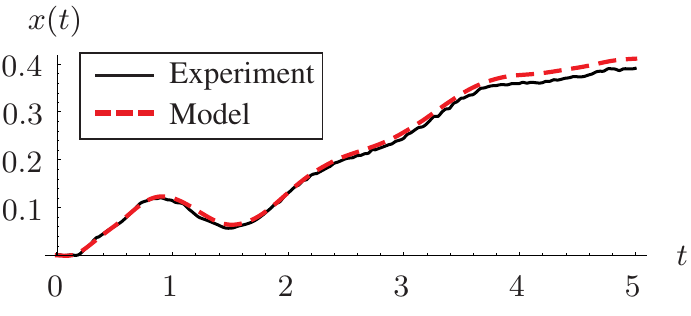}}\\
\vspace{-0.2in}
\subfloat[Link 1 angle, $\phi_1(t)$]{
\hspace{-0.35in}
\includegraphics{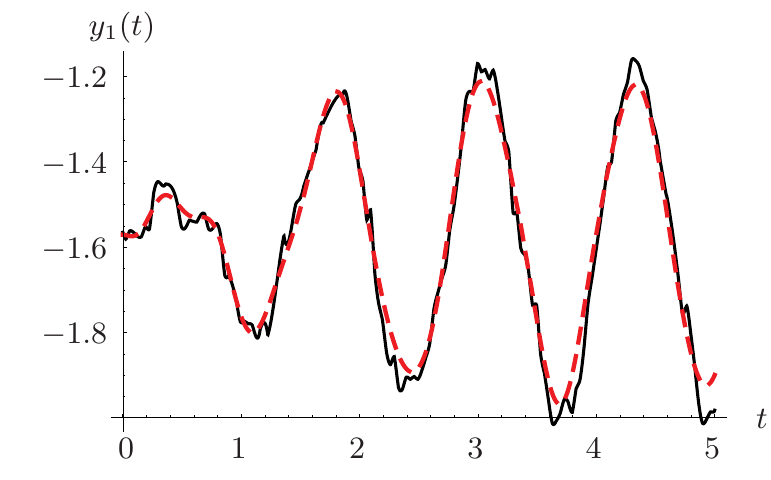}}\\
\vspace{-0.17in}
\subfloat[Link 2 angle, $\phi_2(t)$]{
\hspace{-0.35in}
\includegraphics{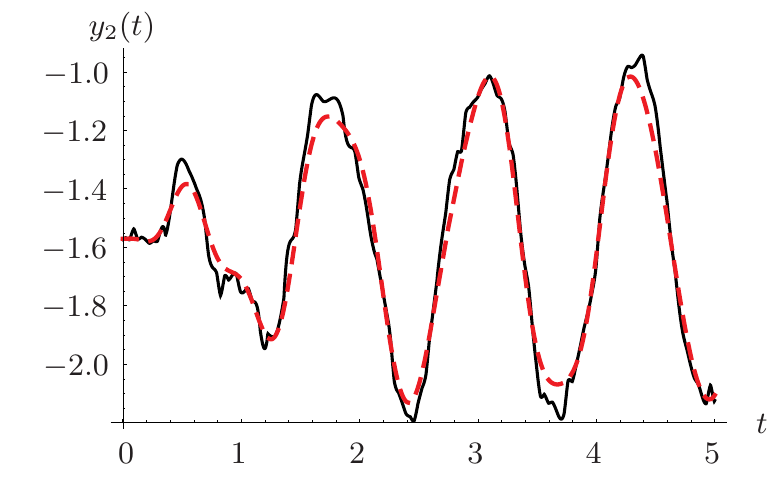}}\\
\caption{Comparisons of the measured trajectories and estimated model for the optimized trajectory based on the Fisher information metric.}
\label{fig:optexp}
\end{figure}

\subsection{Cramer-Rao Lower Bound}
The estimator's covariance result should also be compared with the theoretical Cramer-Rao bound.  As discussed in Section II, the Cramer-Rao bound places an absolute lower bound on the variance of the parameter estimate that can be obtained using the batch least-squares estimator or other unbiased estimator.  The bound is given in (\ref{eq:cramerrao}). 

Table \ref{tab:rao} lists the Cramer-Rao bounds for the initial and optimized trajectories.  The covariance of the initial trajectory is clearly subject to a higher bound than that of the optimized trajectory.  Due to round-off and other numerical errors in the algorithms and Monte-Carlo simulations, the covariance of the Monte-Carlo estimates is higher than the lower bound; however, overall remains within a factor of 3 of the predicted best-case variance estimates according to the Cramer-Rao bound.

\begin{table}[t]
\caption{Experimental Results}
\label{results}
\centering
\begin{tabular}{|l c c|}
\hline
&$m_1$ (kg)&$c$ (g/sec)\\
Initial Estimate:&0.085&0.500\\
Measured Actual Value:&0.110&0.180\\
&&\\
Initial Trajectory Estimate:&0.085&0.500\\
\% Error from Actual Value:&22.7\%&316.7\%\\
&&\\
Optimal Trajectory Estimate:&0.107&0.211\\
\% Error from Actual Value:&2.7\%&17.2\%\\
\hline
\end{tabular}
\end{table}

\begin{figure*}[t]
\centering

\subfloat[Initial trajectory]{
\includegraphics{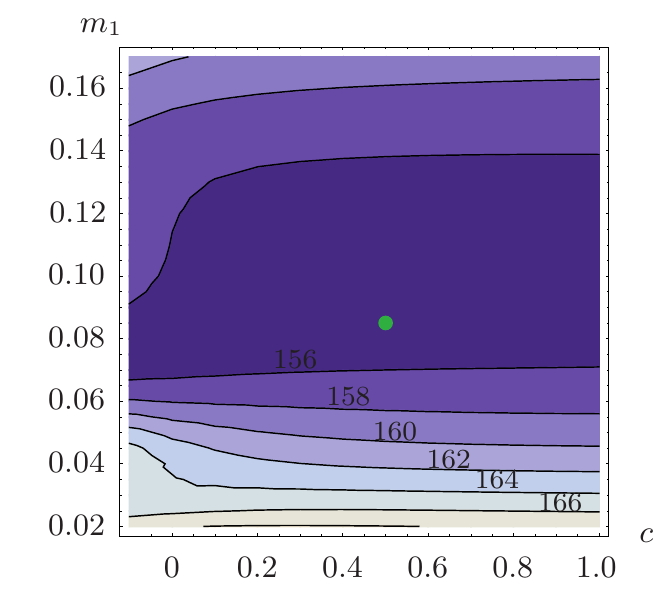}}
\hspace{0.5in}
\subfloat[Optimized trajectory]{
\includegraphics{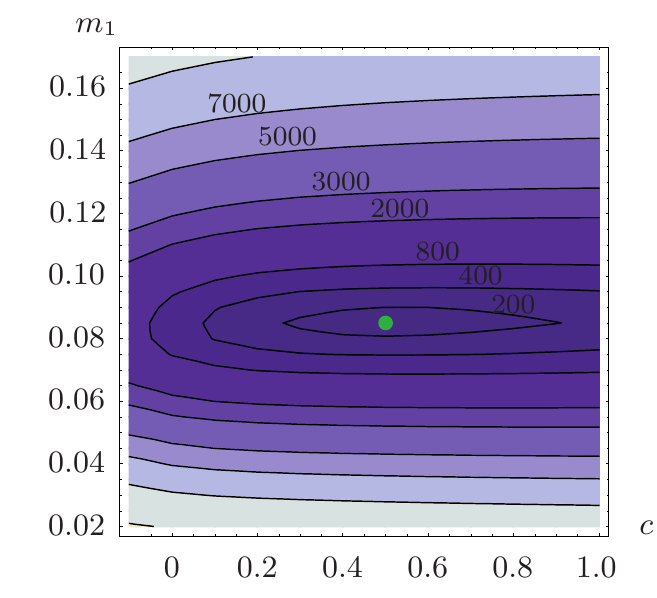}}\\
\caption{Comparisons of parameter optimization cost contours with isolines indicating the estimator cost, $\beta(\theta)$. The deterministic estimate of the parameter is shown by the green dot in the contour plot.  The parametric map of the optimized trajectory is significantly more convex than the initial trajectory.}
\label{fig:optcontour}
\end{figure*}

\subsection{Experimental Results}

After validating the optimization results in simulation, the trajectories were tested on the experimental testbed to determine their experimental effectiveness.  Each trajectory was run on the system with the observed angles and position of the robot recorded by the Kinect tracking system.  The recorded data can be seen in Figs. \ref{fig:initexp} and \ref{fig:optexp}.

Using the collected data as the reference for the least-squares parameter optimization, the parameter set estimate from (\ref{eq:initialestimate}) is used as the initial estimate for the parameter optimization routine.  Using the batch-least squares estimation method, the best estimates of the parameters were found based on the data collected.  The results can be seen in Table \ref{results}.  Actual values of the parameters for the experimental system were obtained by disassembling the pendulum system.  The mass of each link was determined by individually weighing each link, and the damping coefficient was obtained from a batch-least squares estimate of a single link in a free swinging trajectory.

Since the noise present in the initial trajectory is so high relative to the observed angles, the parameter estimation algorithm is not able to make any progress from the initial estimate of the parameters.  This results in a $22.7 \%$ error in the mass and $316.7 \%$ error in the damping coefficient from the baseline values.  The optimized trajectory provides parameter estimates with only $2.7\%$ error in the mass and $17.2\%$ error in the damping coefficient values.

A contour plot of the parameter estimator cost $\beta$ is shown in Fig. \ref{fig:optcontour}.  Using the experimental trajectory data, the cost was computed for a portion of the parameter space.  The figure shows a well-defined optimization basin for the optimized trajectory where the basin of the initial experimental trajectory is far less convex.  This optimized contour plot illustrates why the data from the optimized trajectory provides the better estimate of the parameters of interest given the observed measurements.  The E-optimality cost function, however, does not explicitly condition the optimized basin as seen in Fig. \ref{fig:optcontour}b.  In some cases, it may be advantageous to use an additional approach to improve the condition number of the optimization problem as well \cite{Wilson2013}. 

It is important to note that the error in the experimental case is far greater than the predicted Cramer-Rao bound in Table III.  This is due to additional disturbances and unmodeled effects such as out of plane swinging motion, error in the control signal, and sensor nonlinearities.  These effects will cause bias error in the estimator; however, even with these sources of bias, the optimized trajectory provides a much better estimate of model parameters than the initial system trajectory.

\section{Conclusion}
This paper presented a method to optimize the trajectory of a nonlinear system, maximizing the Fisher information with respect to a set of model parameters.    
The simulation results of the cart-pendulum simulation show that the optimization algorithm results in an increase of the minimum eigenvalue of the Fisher information as well as a decrease in the estimation variance from Monte-Carlo simulation.  Additionally, experimental results confirm a substantial improvement in parameter estimation accuracy when using the optimized trajectory.  

While the formulation of the algorithm allows for general, nonlinear dynamics, limitations on tractable noise models remains a formidable challenge.  Since process noise is assumed to be negligible for the purpose of this algorithm, results may only be useful on systems with accurate input models. Although the algorithm should improve prediction on most systems, predicted precision improvements in simulation may be overestimated compared to the experimental results due to unmodeled bias and system modeling errors.  Still, with appropriate selection of weighting matrices on controls and tracking error, improvement in estimator performance is expected compared to arbitrary choices of the experimental trajectory.

\section*{Acknowledgment}
This material is based upon work supported by the National Institute of Health under NIH R01 EB011615 and by the National Science Foundation
under Grant CMMI 1334609. Any opinions, findings, and conclusions or recommendations expressed in this material are those of the author(s) and do not necessarily reflect the views of the National Science Foundation.

\appendices
\section{Computing the Descent Direction: $\zeta_k(t)$}
\label{app:descent}
To find a descent direction for the optimal control algorithm, (\ref{eq:descent}) must be solved.  As shown in (\ref{eq:descent}), the descent direction depends on the linearization of the cost function, $DJ(P(\xi_k(t)))$, and the local quadratic model, $\langle\zeta_k(t),\zeta_k(t)\rangle$.  Using a quadratic model and expanding the linearizations of the cost function, (\ref{eq:descent}) is rewritten as 
\begin{align}
\arg\min_{\zeta_k(t)}=&\int\limits_{t_0}^{t_f}a(t)^T\bar{z}(t)+b(t)^Tv(t)+\frac{1}{2}\bar{z}(t)^TQ_n\bar{z}(t)\nonumber\\
&+\frac{1}{2}v(t)^TR_nv(t)\hspace{0.05in}dt \label{eq-localquadraticoptimization}
\end{align}  
such that
\begin{equation*}
\dot{\bar{z}}=A\bar{z}+Bv
\end{equation*}
where $a(t)$ and $b(t)$ are the linearizations of the cost function with respect to $\bar{x}$ and $u$, and $Q_n$ and $R_n$ are weighting matrices for the local quadratic model approximation.  Design of these weighting matrices can lead to faster convergence of the optimal control algorithm depending on the specific problem.  The derivations of the cost function linearizations and the dynamics linearizations are now presented.

\subsection{Cost Function Linearization}
The linearization of the cost function $DJ(P(\xi_k(t)))$ is found by taking the directional derivative of (\ref{eq:trajcostcont}) with respect to the extended states $\bar{x}(t)$ and the control vector $u(t)$.

The derivative of (\ref{eq:trajcostcont}) with respect to $\bar{x}$ yields
\begin{align}
\frac{\partial J}{\partial \bar{x}} = &-\frac{Q_p}{\lambda_{min}^2}\frac{\partial \lambda_{min}}{\partial \bar{x}}\nonumber\\ &+\int\limits_{t_0}^{t_f}\left[(x(t)-x_d(t))^T\cdot Q_\tau\right] dt.
\label{eq:costlin}
\end{align}

Since the cost function involves  eigenvalues of $\tilde{I}(\theta)$, this linearization requires the calculation of the derivatives of eigenvalues.  A process for this calculation was formalized by Nelson \cite{Nelson1976}.  Given an eigensystem of the form
\begin{equation*}
AX = X\Lambda
\end{equation*}
where $\Lambda$ is a diagonal matrix of distinct eigenvalues $(\lambda_1,\lambda_2,...\lambda_n)$, and $X$ is the associated matrix of eigenvectors, the derivative of an eigenvalue $\lambda_m$ is given by

\begin{equation*}
D_x\lambda_m=\omega_m^T\cdot D_xA \cdot \nu_m
\end{equation*}
where $\omega_m$ is the left eigenvector and $\nu_m$ is the right eigenvector associated with $\lambda_m$.

If eigenvalues are not distinct, ie. multiplicity greater than 1, the Nelson's method does not hold since the choice of eigenvectors is not unique.  However, in the case of repeated eigenvalues and repeated eigenvalue derivatives, a set of eigenvectors can be determined up to a scalar multiplier as detailed in \cite{Friswell1996}.  Once the eigenvectors are calculated, each eigenvalue and eigenvector pair can be used to compute the direction of steepest descent for the objective.  

Using these methods to compute the eigenvalue derivative, $\frac{\partial \lambda_{min}}{\partial \bar{x}}$ from (\ref{eq:costlin}) can be calculated.  Taking the derivative of the eigenvalue of $\tilde{I}(\theta)$ from (\ref{eq:continfo}) with respect to the extended state yields
\begin{equation*}
\frac{\partial \lambda_{s}}{\partial \bar{x}}=\omega_{s}^T \frac{\partial}{\partial \bar{x}}\left(\int_{t_0}^{t_f}\Gamma_\theta(t)^T\cdot\Sigma^{-1}\cdot\Gamma_\theta(t)\hspace{0.04in}dt\right)\nu_s
\end{equation*}
where $s$ denotes the index of the minimum eigenvalue and eigenvector.  Since the partial derivative and eigenvectors are evaluated only at the final time, the equation can be rewritten to a running cost formulation given by
\begin{equation}
\frac{\partial \lambda_{s}}{\partial \bar{x}}=\int_{t_0}^{t_f}\omega_{s}^T \frac{\partial}{\partial \bar{x}}\left(\Gamma_\theta(t)^T\cdot\Sigma^{-1}\cdot\Gamma_\theta(t)\right)\nu_s \hspace{0.05in}dt.
\label{eq:at2}
\end{equation}
Finally, differentiating the inner product of the gradients yields
\begin{align}
\label{at3}
\frac{\partial}{\partial \bar{x}}&\left(\Gamma_\theta(t)^T\cdot\Sigma^{-1}\cdot\Gamma_\theta(t)\right) = \\
&\left[
\begin{array}{c}
2\hspace{0.03in}\Gamma_\theta(t)^T \cdot \Sigma^{-1} \cdot\left(D^2_xg(\cdot)\cdot \psi(\cdot)+D_xD_\theta g(\cdot)\right)\\
2\hspace{0.03in}\Gamma_\theta(t)^T \cdot \Sigma^{-1} \cdot D_x g(\cdot)\cdot E
\end{array}
\right]
\nonumber
\end{align}
where $E$ is a tensor of the form
\begin{equation*}
E_{ijkl} = \delta_{ik}\delta_{jl}
\end{equation*}
with $\delta$ as the Kronecker delta function.

Combining equations (\ref{eq:costlin}), (\ref{eq:at2}), and (\ref{at3}), $a(t)$ in (\ref{eq-localquadraticoptimization}) is given by,
\begin{align*}
a(t&)=\left[\begin{array}{c}
(x(t)-x_d(t))^T Q_\tau\\
\{0\}^{1\times n\times p}
\end{array}\right]-\\
&\frac{Q_p}{\lambda_{min}^2}\omega_s^T\left[
\begin{array}{c}
2\hspace{0.03in}\Gamma_\theta(t)^T \Sigma^{-1} \left(D^2_xg(\cdot) \psi(t)+D_xD_\theta g(\cdot)\right)\\
2\hspace{0.03in}\Gamma_\theta(t)^T  \Sigma^{-1}  D_x g(\cdot) E
\end{array}
\right]\nu_s.
\end{align*}

The linearization $b(t)$ from (\ref{eq-localquadraticoptimization}), defining the derivative of the cost function with respect to the controls $u(t)$, is given by
\begin{equation*}
b(t)= u(t)^T\cdot R_\tau
\end{equation*}
where $R_{\tau}$ is the weighting matrix from (\ref{eq:trajcostcont}).

\subsection{Dynamics Linearization}
The two other quantities needed to compute the descent direction are $A(t)$ and $B(t)$ -- the linearizations of the dynamics.  The descent direction $\zeta_k$ will satisfy the linear constraint ODE given by
\begin{equation*}
\dot{\bar{z}}_k(t) = A(t)\bar{z}_k(t)+B(t)v_k(t)
\end{equation*}
where $A(t)$ is the linearization of the nonlinear dynamics given by (\ref{eq:dynamics}) and (\ref{eq:gradient}) with respect to $\bar{x}(t)$, and $B(t)$ is the linearization with respect to $u(t)$.
The linearization $A(t)$ of the dynamics with respect to the extended state $\bar{x}(t)$ is given by
\begin{eqnarray*}
A(t)=& \left[\begin{array}{cc}
\frac{\partial \dot{x}}{\partial x}&\frac{\partial \dot{x}}{\partial \psi}\\
\frac{\partial \dot{\psi}}{\partial x}&\frac{\partial \dot{\psi}}{\partial \psi}
\end{array}\right]\\
=&\left[\begin{array}{cc}
D_xf(\cdot)&\{0\}^{n\times n\times p}\\
D^2_xf(\cdot)\cdot\psi(t)+D_xD_\theta f(\cdot)&D_xf(\cdot)\cdot E
\end{array}\right].
\end{eqnarray*}

In addition to the state linearizations, the control linearizations are required.  This linearization matrix $B(t)$ is given by
\begin{eqnarray*}
B(t)=& \left[\begin{array}{c}
\frac{\partial \dot{x}}{\partial u}\vspace{0.02in}\\
\frac{\partial \dot{\psi}}{\partial u}
\end{array}\right]\\
\nonumber
=& \left[\begin{array}{c}
D_uf(\cdot)\\
D_uD_xf(\cdot)\cdot\psi(t)+D_uD_\theta f(\cdot)
\end{array}\right].
\end{eqnarray*} 

Given the linearizations $a(t), b(t), A(t),$ and $B(t)$,  (\ref{eq-localquadraticoptimization}) can be used to compute the descent direction $\zeta_k(t)$.
At each iteration of the optimization algorithm, the perturbed trajectory $\eta_k(t)+\gamma_k\zeta_k$ must be projected to satisfy the dynamics.  As shown in Algorithm 2, the process is repeated until a termination criteria is satisfied.

\bibliographystyle{ieeetr}
\IEEEtriggeratref{16}
\bibliography{tro_fim}

\vfill
\end{document}